\documentclass[conference]{IEEEtran}
\IEEEoverridecommandlockouts

\usepackage{cite}
\usepackage{tikz}
\usepackage{float}
\usetikzlibrary{arrows.meta, positioning}
\usepackage[none]{hyphenat}
\usepackage{amsmath,amssymb,amsfonts}

\definecolor{HeaderPink}{RGB}{190,65,110} \definecolor{RowPink}{RGB}{252,230,239}

\usepackage{algorithmic}
\usepackage{graphicx}
\usepackage{booktabs}
\usepackage{array}
\usepackage{amssymb}
\usepackage{hyperref}
\usepackage{tabularx}
\usepackage{array}
\usepackage{xcolor}
\usepackage[table]{xcolor}
\usepackage{booktabs}

\definecolor{HeaderOrange}{RGB}{230,120,20}
\definecolor{RowOrange}{RGB}{255,242,230}
\usepackage{booktabs}
\usepackage{tabularx}
\usepackage{array}
\usepackage[table]{xcolor}

\definecolor{HeaderBlue}{RGB}{215,230,250}
\definecolor{RowGray}{RGB}{248,248,248}

\newcolumntype{Y}{>{\centering\arraybackslash}X}
\usepackage[table]{xcolor}
\usepackage{multirow}
\usepackage{booktabs}
\usepackage{tabularx}
\usepackage{array}
\usepackage[table]{xcolor}

\definecolor{HeaderBlue}{RGB}{215,230,250}
\definecolor{RowGray}{RGB}{248,248,248}

\newcolumntype{Y}{>{\centering\arraybackslash}X}
\usepackage{pifont}
\usepackage{tabularx}
\usepackage{booktabs}
\usepackage{colortbl}
\usepackage{courier}
\usepackage{textcomp}
\def\BibTeX{{\rm B\kern-.05em{\sc i\kern-.025em b}\kern-.08em
    T\kern-.1667em\lower.7ex\hbox{E}\kern-.125emX}}

\begin{document}

\title{IDRAAK: From Multi-Agent NLP to Few-Shot Prompting for Semantic Drift Detection in Technical Requirements  

}

\author{
\IEEEauthorblockN{Shiva Ahir}
\IEEEauthorblockA{
\textit{Department of Electrical \& Computer Engineering} \\
\textit{Stony Brook University} \\
New York, United States of America \\
shiva.ahir@stonybrook.edu
}
}
\maketitle

\begin{abstract}
Translating technical requirements across languages can introduce semantic drift, altering numerical constraints, polarities, modalities, or other specification-critical meaning. IDRAAK is presented as an interpretable framework for detecting such drift using a language-independent Semantic Requirement Representation (SRR), with six detection workflows evaluated, ranging from deterministic comparison to multi-agent verification and few-shot prompting. On 890 synthetic perturbations across 300 requirements from 10 engineering domains, a single LLM call with six few-shot examples achieves MCC=0.888 and F1=0.983, outperforming the evaluated structured and multi-stage alternatives. Further evaluation on PAWS-X~\cite{yang2019pawsx} (805 pairs, 5 languages) and XNLI~\cite{conneau2018xnli} (700 pairs, 7 languages) exposes complementary strengths and limitations of structured and LLM-based approaches. Deterministic SRR comparison performs strongly on technical requirements (F1=0.898) but poorly on general-domain text (F1=0.012), while structured evidence improves performance on adversarial paraphrases. Post-hoc Platt scaling further improves confidence calibration. The results demonstrate that increased agentic complexity does not necessarily improve semantic-drift detection and that simple few-shot prompting can provide a strong and efficient alternative.



{\small
\checkmark\ \textbf{Open Source:} \url{https://github.com/shivaahir158/idraak}

}

\end{abstract}



 \section{Introduction}
Technical requirements define precise specifications governing safety-critical systems across aerospace, automotive, medical, and industrial domains
\cite{pohl2010}. When translated across languages, these requirements risk semantic drift: subtle meaning shifts that alter numerical constraints, invert
polarities, weaken modalities, or restructure conditional logic while preserving surface fluency \cite{berry2003}. Unlike general translation errors, such
drift can introduce catastrophic specification faults, as when \textit{`shall respond within 50 milliseconds''} becomes \textit{`should respond within 50 microseconds.''}

\begin{figure}[t]
\centering
\includegraphics[width=\columnwidth]{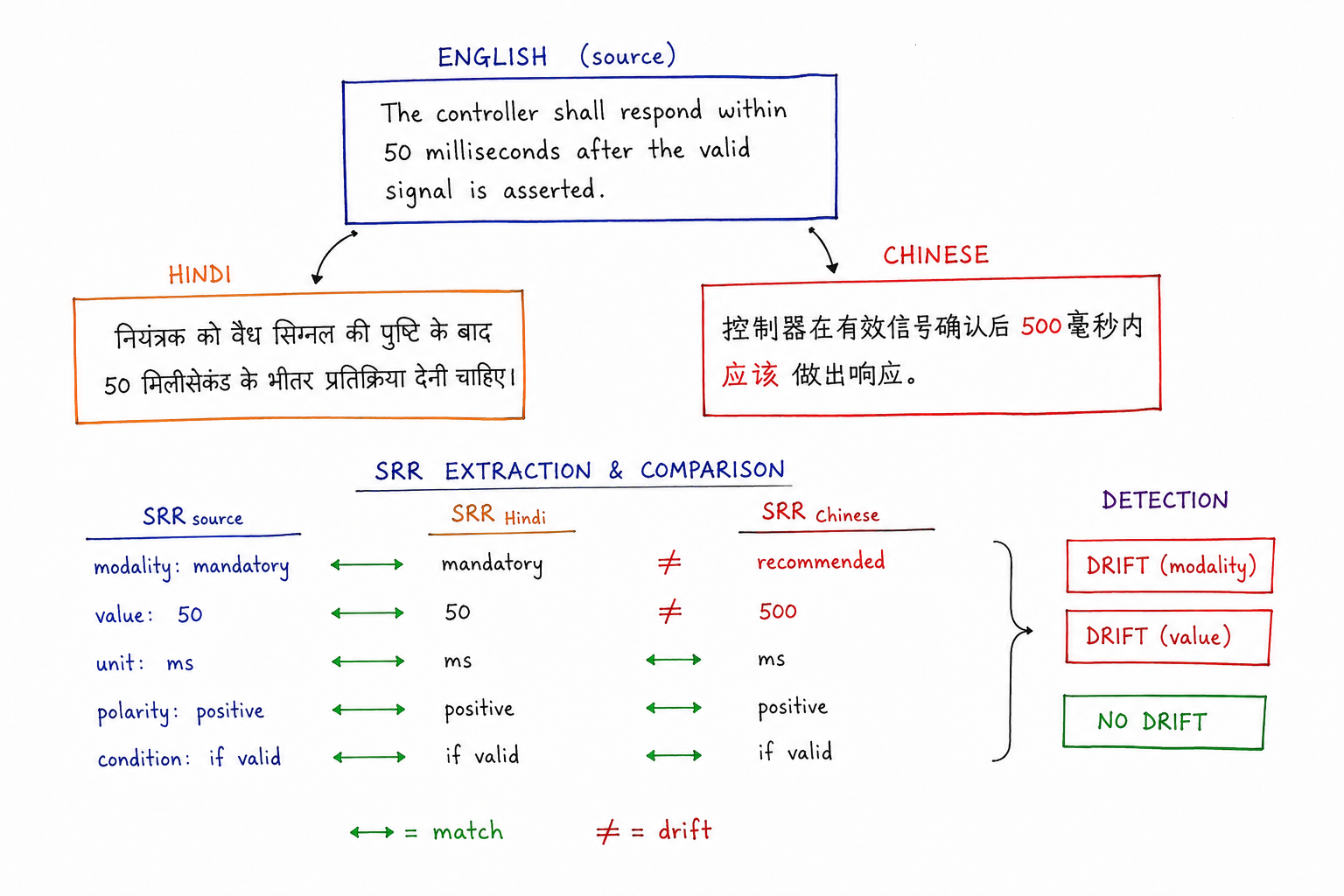}
\caption{Cross-lingual semantic drift detection through structured requirement comparison. The Hindi translation preserves the source semantics, while the Chinese translation introduces numerical and modality drift.}
\label{fig:crosslingual_drift}
\end{figure}

Figure~\ref{fig:crosslingual_drift} illustrates this problem across
languages: a semantically faithful translation preserves the underlying
technical constraints despite lexical differences, whereas a drifting
translation may alter numerical values or modality while remaining
grammatically fluent.

Existing translation quality metrics are inadequate for this task. BLEU \cite{papineni2002} measures surface overlap and is insensitive to meaning-altering
substitutions. Neural metrics such as BERTScore \cite{zhang2020bertscore} and COMET \cite{rei2020} capture semantic similarity but produce opaque scalar
scores without identifying what changed or why. Cross-lingual embeddings \cite{conneau2020unsupervised} similarly lack the structured interpretability
required for engineering review and certification workflows.

Large language models offer new possibilities for semantic evaluation \cite{openai2023gpt4, brown2020}, and multi-agent architectures have been proposed to
improve reliability through specialized agent collaboration \cite{wu2023autogen, du2024}. However, whether multi-agent complexity improves drift detection
over simpler prompting approaches remains empirically unexamined.

This paper presents IDRAAK (\textit{Interpretable Drift Recognition with Agent-Augmented Knowledge}), an interpretable framework for detecting semantic drift in multilingual technical requirements. The contributions are:

\begin{enumerate}
\item A \textbf{Semantic Requirement Representation (SRR)} that decomposes requirements into language-independent typed components including numerical
constraints, modalities, conditions, temporal expressions, and domain entities, enabling deterministic field-level comparison across 15 drift categories.

\item A \textbf{systematic comparison of six detection workflows} spanning deterministic comparison, single-agent hybrid extraction, few-shot direct
prompting, ensemble methods combining structured evidence with LLM judgment, full multi-agent verification with eight specialized agents, and
back-translation.

\item Empirical evidence that a single LLM call
with six few-shot examples (MCC=0.888)
substantially outperforms the evaluated structured
and multi-stage configurations on 890 perturbations
across 300 requirements from 10 engineering domains,
demonstrating that increased architectural complexity
does not necessarily improve drift detection. Further validation on PAWSX \cite{yang2019pawsx} and XNLI \cite{conneau2018xnli} benchmarks, demonstrating cross-domain generalization.
\end{enumerate}

\section{Related Work}

\subsection{NLP for Requirements Engineering}

Natural language processing has been extensively applied to requirements engineering tasks including ambiguity detection, consistency checking, and
automated classification \cite{zhao2021, ferrari2017}. Berry et al. \cite{berry2003} catalogued linguistic sources of ambiguity in software specifications,
establishing foundational categories that inform drift taxonomies. However, most work focuses on monolingual English requirements. Cross-lingual
requirements analysis remains underexplored, particularly for detecting meaning shifts introduced during translation rather than inherent specification
defects.

\subsection{Translation Quality Estimation}

Traditional translation evaluation relies on reference-based metrics. BLEU \cite{papineni2002} measures n-gram overlap but is insensitive to semantically
significant substitutions that share surface similarity. Neural learned metrics including BERTScore \cite{zhang2020bertscore}, COMET \cite{rei2020}, and
BLEURT \cite{sellam2020} leverage contextual embeddings for improved semantic sensitivity, and recent WMT evaluations have demonstrated their superiority
over surface metrics \cite{freitag2022}. However, these metrics produce scalar similarity scores without structured explanations, making them unsuitable for
safety-critical domains where reviewers must understand exactly which technical attributes changed. SRR-based comparison addresses this gap by
producing field-level evidence for each detected drift.

\subsection{Semantic Textual Similarity and Paraphrase Detection}

Semantic textual similarity (STS) has been studied extensively through shared tasks \cite{agirre2012, cer2017} and dedicated benchmarks. PAWS-X
\cite{yang2019pawsx} provides adversarial cross-lingual paraphrase pairs where high lexical overlap coexists with different meanings, making it particularly
relevant for evaluating drift detection robustness. XNLI \cite{conneau2018xnli} extends natural language inference to 15 languages, enabling cross-lingual
evaluation of semantic reasoning. Cross-lingual sentence encoders such as XLM-RoBERTa \cite{conneau2020unsupervised} and multilingual Sentence-BERT
\cite{reimers2019} provide strong baselines for similarity tasks but, like scalar metrics, lack interpretability at the attribute level. This work differs
from standard STS by targeting structured technical content where drift in a single field such as a unit or numerical value constitutes a meaningful
semantic change that aggregate similarity scores may miss.

\subsection{Large Language Models as Evaluators}

LLMs have demonstrated strong performance as text evaluators when provided with appropriate prompts \cite{openai2023gpt4}. Brown et al. \cite{brown2020}
established that few-shot prompting enables task adaptation without fine-tuning, and subsequent work has shown that example selection significantly impacts
in-context learning effectiveness \cite{liu2022, min2022}. Chain-of-thought prompting \cite{wei2022} further improves reasoning on complex tasks. The direct
judge workflow builds on these findings, demonstrating that six carefully selected few-shot examples covering representative drift types are sufficient to
achieve strong detection performance without task-specific training.

\subsection{Multi-Agent LLM Systems}

Multi-agent frameworks decompose complex tasks across specialized LLM-based agents that collaborate through structured interaction. AutoGen
\cite{wu2023autogen} enables flexible multi-agent conversations, MetaGPT \cite{hong2024metagpt} assigns distinct roles to agents for software development,
and multi-agent debate has been shown to improve factuality and reasoning \cite{du2024, chan2024chateval}. The underlying hypothesis is that specialized
agents with focused responsibilities outperform monolithic approaches. This work provides a controlled empirical test of this hypothesis in the context of
semantic drift detection, finding that error propagation across agent boundaries outweighs specialization benefits, and that simpler workflows consistently
achieve superior performance.

\subsection{Confidence Calibration}

Modern neural networks and LLMs are often poorly calibrated, producing confidence scores that do not reflect true correctness probabilities \cite{guo2017,
kadavath2022}. Post-hoc calibration methods including Platt scaling \cite{platt1999}, isotonic regression \cite{zadrozny2002}, and temperature scaling
\cite{guo2017} can improve reliability without retraining. These methods are then applied to drift detection confidence scores and demonstrate that Platt scaling
reduces expected calibration error from 0.452 to 0.013, an important property for downstream decision-making in safety-critical requirement review.

\subsection{Faithfulness in Text Generation}

Related work on faithfulness in abstractive summarization \cite{maynez2020} and hallucination detection \cite{ji2023} shares the concern with meaning
preservation but operates in a generation context rather than translation verification. Back-translation has been studied as both a data augmentation
technique \cite{edunov2018} and a quality indicator for translation faithfulness. Back-translation is incorporated as one of the six workflows, enabling round-trip consistency verification as a complementary signal for semantic-drift detection.


\begin{figure}[t]
\centering
\includegraphics[width=\columnwidth]{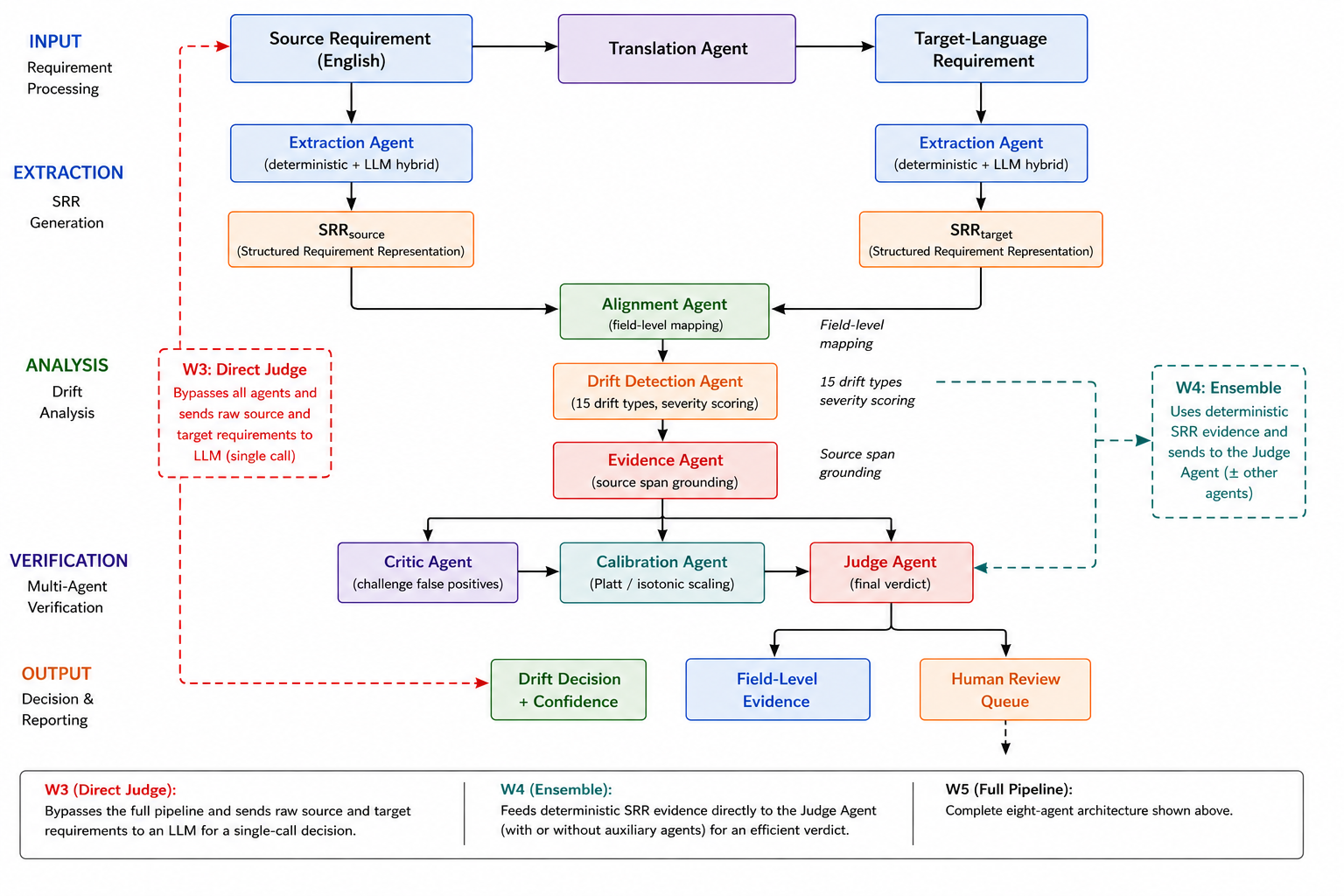}
\caption{Architecture of the \textbf{IDRAAK} framework (W5). The system comprises five phases and eight specialized agents for cross-lingual drift detection. W3 and W4 represent simplified workflow variants.}
\label{fig:architecture}
\end{figure}

  \section{Methodology}                               
  \subsection{Semantic Requirement Representation}                                                                                      
  The core of IDRAAK is the Semantic Requirement Representation (SRR), a language-independent structured form that decomposes a technical requirement into  
  typed components. Given a requirement text $r$, the extraction function $\mathcal{E}(r) \rightarrow S$ produces an SRR $S$ containing:

  \begin{itemize}
      \item \textbf{Actor, Action, Object}: the subject performing an action on a target
      \item \textbf{Modality}: obligation level (mandatory, recommended, permitted, optional, forbidden)
      \item \textbf{Polarity}: positive or negative assertion
      \item \textbf{Numerical constraints}: parameter-operator-value-unit tuples (e.g., \texttt{latency $\leq$ 50 ms})
      \item \textbf{Conditions}: typed clauses (if, when, unless, only\_if) with optional nesting
      \item \textbf{Temporal constraints}: timing relations (before, after, within) with values and reference events
      \item \textbf{Ordering constraints}: sequencing between events (first, second, strict/non-strict)
      \item \textbf{Exceptions}: exclusion clauses (unless, except, provided\_that)
      \item \textbf{Entities and Relations}: named domain objects with roles and typed predicates
  \end{itemize}

  All components are defined as Pydantic v2 models \cite{pydantic} with strict type validation. The representation is designed so that two requirements with
  identical meaning in different languages produce structurally equivalent SRRs.

  \subsection{SRR Extraction}

  Three extraction strategies are implemented. \textbf{Deterministic extraction} uses regular expressions and pattern matching to parse modality markers, numerical values with units, conditional keywords, temporal expressions, and polarity indicators. It is fast and reproducible but limited to patterns
  explicitly encoded. \textbf{LLM extraction} prompts an LLM to produce a structured SRR as JSON, capturing semantic content that resists pattern matching.
  \textbf{Hybrid extraction} runs both methods and merges results using a priority strategy where deterministic values always override LLM outputs for
  comparison-critical fields (numerical constraints, modality, polarity, units), while LLM outputs fill fields where deterministic extraction yields nothing
  (entities, relations, qualifiers). This merge strategy proved critical: without it, LLM extraction introduced noise into fields where deterministic parsing
  was already accurate.

  \subsection{Drift Detection}

  Given a source requirement $r_s$ and candidate translation $r_c$, drift detection proceeds by extracting SRRs $S_s = \mathcal{E}(r_s)$ and $S_c =
  \mathcal{E}(r_c)$, then computing field-level differences. The comparison engine performs unit-aware numerical matching (e.g., recognizing that 1 kB = 1024
  bytes), operator compatibility checks, modality ordering comparison, polarity alignment, and set-based matching for entities and conditions. Each difference
   is classified into one of 15 drift types: numerical, unit, polarity, modality, condition, temporal, threshold, entity, relation, exception, omission,
  addition, terminology, scope, and reference. Differences are assigned severity levels (none, low, medium, high, critical) and aggregated into a binary drift
   decision with an associated confidence score.

  \subsection{Detection Workflows}

  Figure~\ref{fig:architecture} provides an overview of the IDRAAK
architecture and the relationship among the evaluated workflows.
The complete W5 configuration executes the full multi-agent pipeline,
while W3 and W4 bypass portions of this pipeline to evaluate whether
simpler decision mechanisms can achieve comparable or superior
semantic-drift detection. Six workflows of increasing complexity are implemented:

  \textbf{(W1) Deterministic comparison.} Extracts SRRs using regex-based parsing and compares fields deterministically. Produces interpretable field-level
  evidence but is limited to patterns present in the extraction rules.

  \textbf{(W2) Structured single-agent.} Uses hybrid extraction (deterministic + LLM) followed by the same field-level comparison. The LLM enriches extraction
   coverage while deterministic priority preserves precision on critical fields.

  \textbf{(W3) Direct judge.} A single LLM call receives both requirement texts along with six few-shot examples \cite{brown2020, liu2022} covering
  representative drift types (paraphrase, numerical drift, polarity inversion, entity swap, modality shift, and omission) and returns a binary drift decision
  with confidence and explanation.

  \textbf{(W4) Ensemble.} Runs deterministic SRR extraction and comparison to produce structured evidence, then passes this evidence alongside the original
  texts to an LLM judge. The LLM makes the final decision informed by both the raw texts and the structured differences.

  \textbf{(W5) Full multi-agent IDRAAK.} Eight specialized agents execute sequentially: a \textit{translation agent} handles language conversion, an
  \textit{extraction agent} produces SRRs, an \textit{alignment agent} maps corresponding fields, a \textit{drift detection agent} identifies differences, an
  \textit{evidence agent} grounds findings in source text spans, a \textit{critic agent} challenges potential false positives, a \textit{calibration agent}
  adjusts confidence scores, and a \textit{judge agent} renders the final verdict. Each agent operates with a deterministic fallback when LLM calls fail.

  \textbf{(W6) Back-translation.} Translates the candidate back to the source language and compares the back-translated text against the original using SRR
  comparison \cite{edunov2018}, providing a round-trip consistency signal.

\subsection{Confidence Calibration}

Raw confidence scores from LLM-based workflows are typically poorly calibrated \cite{guo2017, kadavath2022}. Post-hoc calibration is performed using Platt scaling \cite{platt1999}, which fits a logistic regression on the log-odds of raw confidence scores, and isotonic regression \cite{zadrozny2002}, which learns a monotonic mapping from raw scores to calibrated probabilities. Both methods are fitted on validation data and evaluated using Expected Calibration Error (ECE).

\subsection{Evaluation Metrics}

Accuracy, precision, recall, and F1 score are reported for completeness, while Matthews Correlation Coefficient (MCC) \cite{matthews1975} is used as the primary evaluation metric. MCC accounts for all four quadrants of the confusion matrix and is robust to class imbalance \cite{chicco2020}, making it more informative than F1 for this task, where the drift-positive class dominates. AUROC and ECE are additionally reported to assess discrimination and calibration quality, respectively.

\section{Experiments}                  
                                                                                                                                                              
  \subsection{Datasets}                                                                                                                                       
                                                                       
  \subsubsection{Synthetic Benchmark}                                                                                                                         
                                                                                                                                                            
 A total of 300 technical requirements are generated and uniformly distributed across 10 engineering domains: digital hardware, embedded systems, software systems, networking, cybersecurity, safety-critical systems, data processing, financial systems, healthcare devices, and industrial automation (30 per domain). Each requirement is generated from domain-specific templates with randomized actors, actions, signals, numerical values, units, modalities, conditions, and temporal constraints. A controlled perturbation engine subsequently produces labeled variants for each requirement, yielding 890 total samples: 142 semantically equivalent paraphrases (label 0) and 748 drifted variants (label 1) spanning 11 drift types. The resulting distribution is presented in Table~\ref{tab:drift_dist}. All random seeds are fixed to ensure reproducibility.

\begin{table}[t]
\centering
\caption{Distribution of perturbation types in the synthetic benchmark.}
\label{tab:drift_dist}
\renewcommand{\arraystretch}{1.15}
\setlength{\tabcolsep}{14pt}

\begin{tabular}{@{}lr@{}}
\toprule
\rowcolor{HeaderPink}
\textcolor{white}{\textbf{Drift Type}} &
\textcolor{white}{\textbf{Count}} \\
\midrule

\rowcolor{RowPink}
Polarity drift        & 189 \\
Modality drift        & 179 \\
\rowcolor{RowPink}
Numerical drift       & 104 \\
Unit drift            & 89 \\
\rowcolor{RowPink}
Threshold drift       & 53 \\
Temporal drift        & 36 \\
\rowcolor{RowPink}
Scope drift           & 24 \\
Terminology drift     & 23 \\
\rowcolor{RowPink}
Condition drift       & 21 \\
Omission drift        & 19 \\
\rowcolor{RowPink}
Entity drift          & 11 \\

\midrule
\rowcolor{RowPink}
Paraphrase (no drift) & 142 \\

\midrule
\textbf{Total} & \textbf{890} \\
\bottomrule
\end{tabular}

\end{table}

\subsubsection{PAWSX}

Evaluation is conducted on PAWS-X \cite{yang2019pawsx}, a cross-lingual adversarial paraphrase identification benchmark containing sentence pairs with high lexical overlap but potentially different meanings. Test splits across five languages (en, de, es, fr, zh) are used, totaling 805 pairs (486 paraphrase and 319 non-paraphrase). Paraphrase pairs are mapped to no-drift (label 0), while non-paraphrase pairs are mapped to drift (label 1). This benchmark assesses robustness to adversarial lexical overlap, a challenging scenario in which surface similarity is deliberately misleading.

\subsubsection{XNLI}

Evaluation is conducted on XNLI \cite{conneau2018xnli}, a cross-lingual natural language inference benchmark, using 100 premise-hypothesis pairs per language across seven languages (en, hi, ar, zh, de, fr, es), totaling 700 pairs. Entailment pairs are mapped to no-drift, while contradiction and neutral pairs are mapped to drift. This mapping is imperfect because entailment represents a directional relation, whereas semantic equivalence is symmetric; nevertheless, it provides a useful stress test on non-technical, general-domain text.

  \subsection{Experimental Setup}

  All LLM-based experiments use GPT-4o-mini via the OpenAI API. We evaluate six workflow configurations:

  \begin{enumerate}
      \item \textbf{Structured Single / Deterministic} (W1): regex-based SRR extraction with field-level comparison
      \item \textbf{Structured Single / OpenAI} (W2): hybrid extraction (deterministic + LLM) with field-level comparison
      \item \textbf{Direct Judge / OpenAI} (W3): single LLM call with six few-shot examples
      \item \textbf{Ensemble / OpenAI} (W4): deterministic SRR evidence passed to LLM judge
      \item \textbf{Full IDRAAK / Deterministic} (W5a): eight-agent pipeline with deterministic fallback
      \item \textbf{Full IDRAAK / OpenAI} (W5b): eight-agent pipeline with LLM-powered agents
  \end{enumerate}

  The few-shot examples for W3 were manually selected to cover six representative cases: a semantically equivalent paraphrase, a numerical value change, a
  polarity inversion, an entity substitution, a modality weakening, and an omission. Temperature is set to 0 for all LLM calls to ensure deterministic
  outputs. Each experiment is run on the full dataset without subsampling.

  \subsection{Results}

  \subsubsection{Synthetic Benchmark}

  Table~\ref{tab:synthetic} presents the results on the 890-sample synthetic benchmark, while Figure~3 provides a visual comparison of performance across the evaluated workflows. The direct judge (W3) achieves the strongest overall performance, with an F1 score of 0.983 and an MCC of 0.888, substantially outperforming the other approaches. The structured single-agent workflow with hybrid extraction (W2) achieves an MCC of 0.501, indicating that prioritizing deterministic extraction for comparison-critical fields improves the reliability of the hybrid strategy. Both deterministic baselines (W1 and W5a) obtain identical performance (MCC=0.474), suggesting that decomposing the same deterministic operations across multiple agents provides no additional benefit. The ensemble workflow (W4) achieves an F1 score of 0.926 but a lower MCC of 0.411; however, this result should be interpreted with caution because API rate limiting caused approximately 200 samples to fall back to deterministic processing. Overall, Figure~3 highlights that the comparatively simple few-shot direct judge achieves the best performance despite requiring substantially less architectural complexity than the full multi-agent framework.

\begin{table}[t]
\centering
\caption{Results on the synthetic benchmark (890 perturbations, 300 requirements, 10 domains).}
\label{tab:synthetic}
\renewcommand{\arraystretch}{1.2}

\begin{tabular}{llccc}
\toprule
\rowcolor{HeaderPink}
\textcolor{white}{\textbf{Workflow}} &
\textcolor{white}{\textbf{Provider}} &
\textcolor{white}{\textbf{F1}} &
\textcolor{white}{\textbf{MCC}} &
\textcolor{white}{\textbf{Time}} \\
\midrule

\rowcolor{RowPink}
Structured Single & Deterministic & 0.898 & 0.474 & 0.1s \\
Structured Single & OpenAI        & 0.918 & 0.501 & 181min \\

\rowcolor{RowPink}
\textbf{Direct Judge} & \textbf{OpenAI} &
\textbf{0.983} & \textbf{0.888} & \textbf{17min} \\

Ensemble & OpenAI & 0.926 & 0.411 & 27min \\

\rowcolor{RowPink}
Full IDRAAK & Deterministic & 0.898 & 0.474 & 0.2s \\

\bottomrule
\end{tabular}
\end{table}


  \subsubsection{PAWSX}

  Table~\ref{tab:pawsx} presents results on the PAWSX adversarial benchmark. The ensemble workflow achieves the highest MCC=0.459, slightly outperforming the
  direct judge (MCC=0.451). This reversal from the synthetic benchmark suggests that structured SRR evidence provides a useful inductive bias when the LLM
  must distinguish adversarial pairs with high lexical overlap. Deterministic comparison fails almost completely (F1=0.012) because general-domain sentences
  lack the technical patterns (modalities, numerical constraints, units) that drive SRR-based detection.

\begin{figure}[t]
    \centering
    \includegraphics[width=\columnwidth]{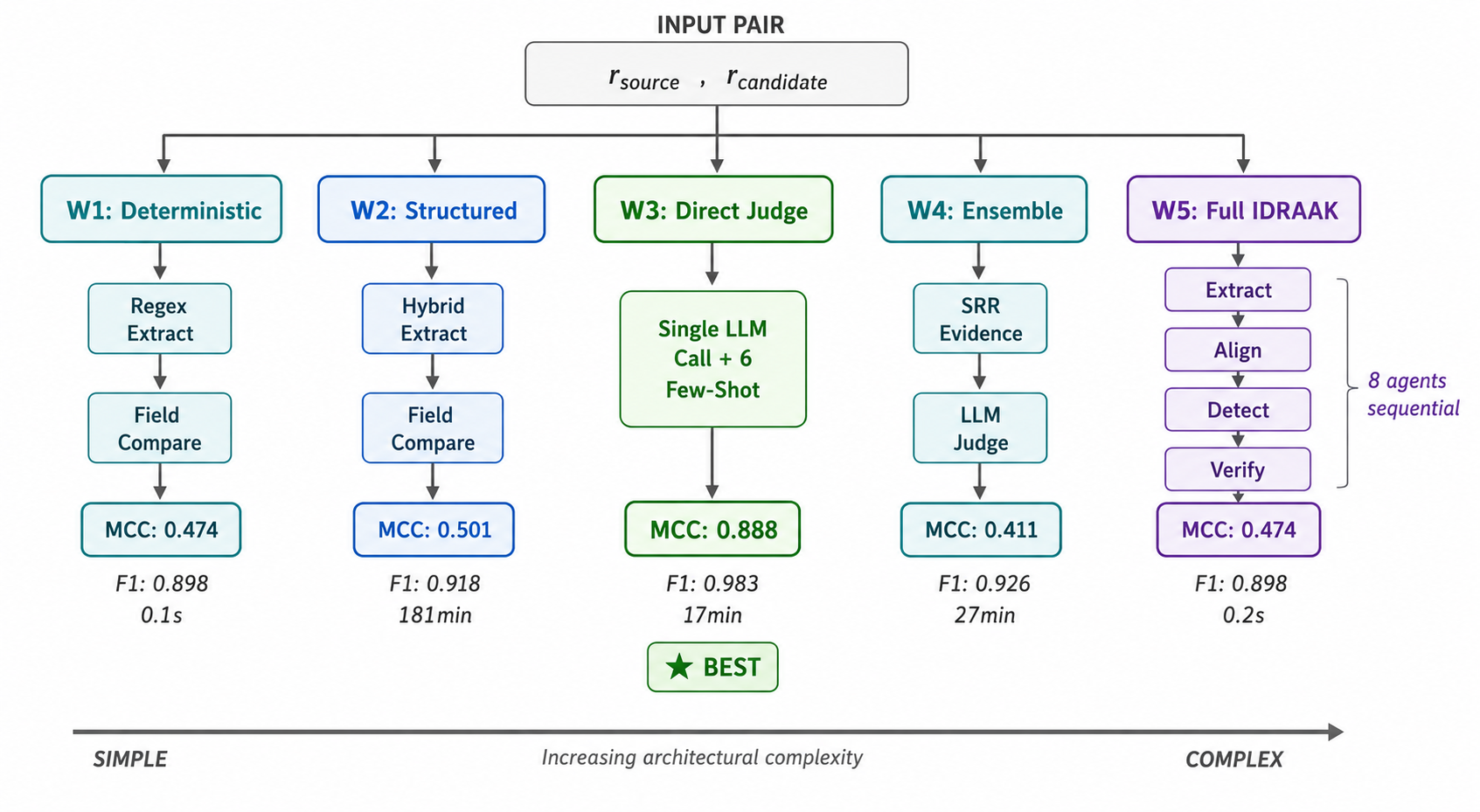}
    \caption{Performance comparison of the five drift-detection workflows. W3 achieves the best overall performance despite its simpler architecture.}
    \label{fig:workflow_comparison}
\end{figure}

\begin{table}[t]
\centering
\caption{Results on PAWSX adversarial paraphrase benchmark (805 pairs, 5 languages).}
\label{tab:pawsx}
\renewcommand{\arraystretch}{1.2}

\begin{tabular}{llccc}
\toprule
\rowcolor{HeaderPink}
\textcolor{white}{\textbf{Workflow}} &
\textcolor{white}{\textbf{Provider}} &
\textcolor{white}{\textbf{F1}} &
\textcolor{white}{\textbf{Acc.}} &
\textcolor{white}{\textbf{MCC}} \\
\midrule

\rowcolor{RowPink}
Structured Single & Deterministic & 0.012 & 0.388 & $-$0.098 \\

Direct Judge & OpenAI & 0.814 & 0.739 & 0.451 \\

\rowcolor{RowPink}
\textbf{Ensemble} & \textbf{OpenAI} &
\textbf{0.817} & \textbf{0.738} & \textbf{0.459} \\

Structured Single & OpenAI & 0.723 & 0.583 & $-$0.003 \\

\rowcolor{RowPink}
Full IDRAAK & Deterministic & 0.012 & 0.388 & $-$0.098 \\

\bottomrule
\end{tabular}
\end{table}

  \subsubsection{XNLI}

  Table~\ref{tab:xnli} presents results on XNLI. All methods achieve low MCC (0.074--0.101), reflecting the fundamental mismatch between entailment and
  semantic drift. LLM-based workflows predict drift for nearly all pairs (343/700 false positives for the direct judge, zero false negatives), indicating that
   the model recognizes non-equivalence in premise-hypothesis pairs but cannot distinguish directional entailment from bilateral drift. Per-language analysis
  shows consistent performance across all seven languages with no significant degradation for any individual language.

\begin{table}[t]
\centering
\caption{Results on XNLI benchmark (700 pairs, 7 languages).}
\label{tab:xnli}
\renewcommand{\arraystretch}{1.2}

\begin{tabular}{llccc}
\toprule
\rowcolor{HeaderPink}
\textcolor{white}{\textbf{Workflow}} &
\textcolor{white}{\textbf{Provider}} &
\textcolor{white}{\textbf{F1}} &
\textcolor{white}{\textbf{Acc.}} &
\textcolor{white}{\textbf{MCC}} \\
\midrule

\rowcolor{RowPink}
Structured Single & Deterministic & 0.396 & 0.533 & 0.074 \\

\textbf{Direct Judge} & \textbf{OpenAI} &
\textbf{0.671} & \textbf{0.510} & \textbf{0.101} \\

\rowcolor{RowPink}
Ensemble & OpenAI & 0.670 & 0.507 & 0.085 \\

Structured Single & OpenAI & 0.669 & 0.520 & 0.091 \\

\rowcolor{RowPink}
Full IDRAAK & Deterministic & 0.396 & 0.533 & 0.074 \\

\bottomrule
\end{tabular}
\end{table}

  \subsubsection{Confidence Calibration}

  Table~\ref{tab:calibration} reports calibration results on XNLI. Raw LLM confidence scores are poorly calibrated across all workflows, with ECE ranging from
   0.120 to 0.464. Platt scaling \cite{platt1999} reduces ECE to below 0.02 in all cases, and isotonic regression \cite{zadrozny2002} achieves near-perfect
  calibration. These results confirm that post-hoc calibration is essential for reliable confidence scores in drift detection, particularly when confidence
  thresholds are used for human review routing in safety-critical workflows.

\begin{table}[t]
\centering
\caption{Calibration results on XNLI (Expected Calibration Error).}
\label{tab:calibration}
\renewcommand{\arraystretch}{1.2}

\begin{tabular}{lccc}
\toprule
\rowcolor{HeaderPink}
\textcolor{white}{\textbf{Workflow}} &
\textcolor{white}{\textbf{Raw ECE}} &
\textcolor{white}{\textbf{Platt}} &
\textcolor{white}{\textbf{Isotonic}} \\
\midrule

\rowcolor{RowPink}
Direct Judge / OpenAI & 0.452 & 0.013 & 0.000 \\

Ensemble / OpenAI & 0.464 & 0.001 & 0.000 \\

\rowcolor{RowPink}
Structured Single / OpenAI & 0.225 & 0.018 & 0.000 \\

Structured Single / Det. & 0.120 & 0.002 & 0.000 \\

\rowcolor{RowPink}
Full IDRAAK / Det. & 0.285 & 0.003 & 0.000 \\

\bottomrule
\end{tabular}
\end{table}

\subsection{Analysis}                                                                                                                                       
                                                                                                                                                            
  \begin{figure}[t]                                                                                                                                           
  \centering                                                                                                                                                
  \includegraphics[width=\columnwidth]{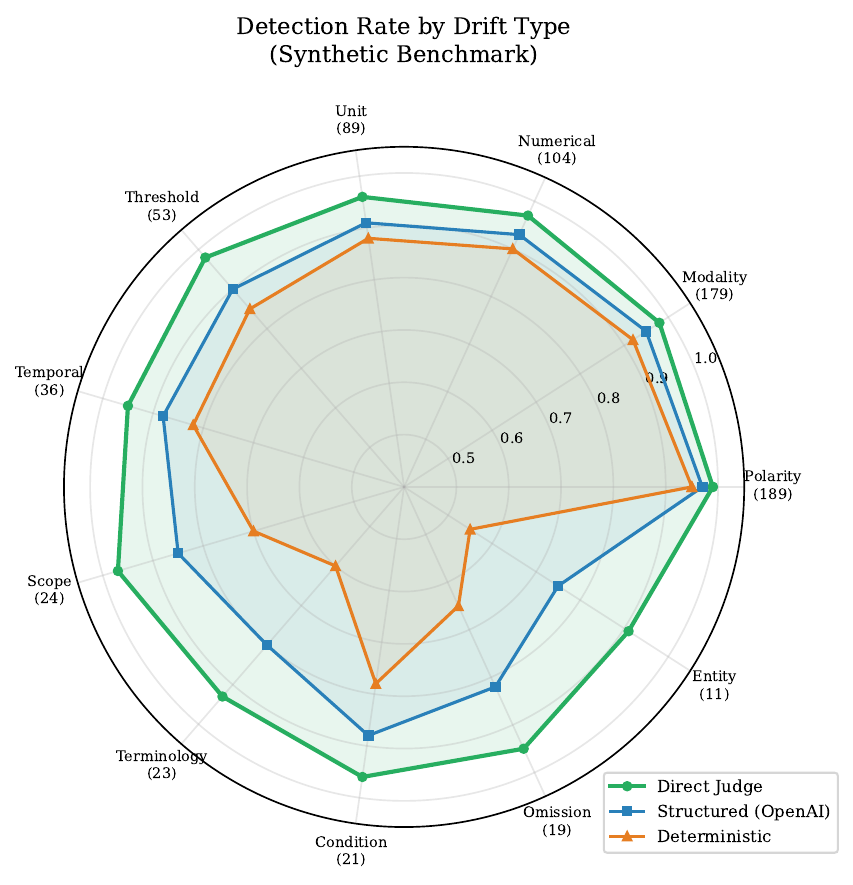}                                                                                               
  \caption{Detection rate by drift type on the synthetic benchmark.}                                                                                        
  \label{fig:drift_radar}
  \end{figure}

  \begin{figure}[t]
  \centering
  \includegraphics[width=\columnwidth]{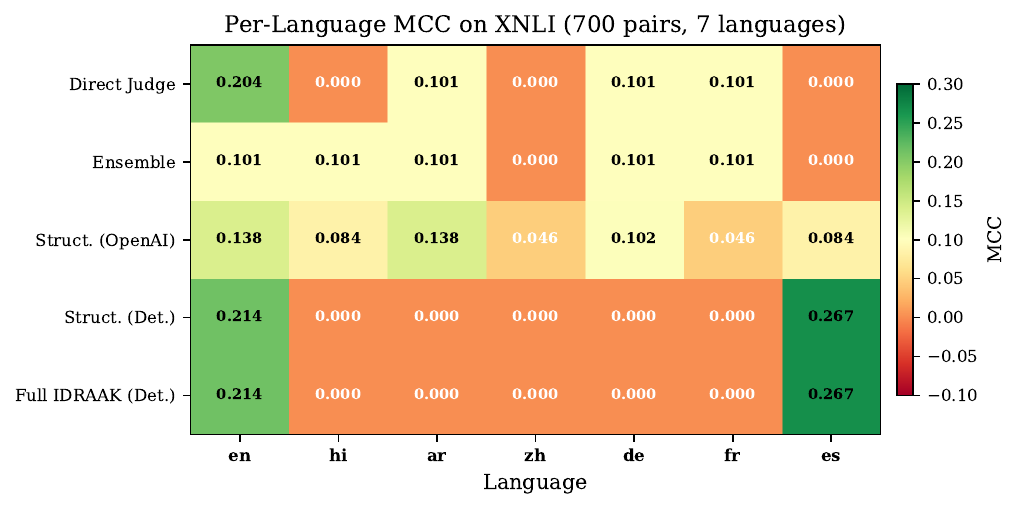}
  \caption{Per-language MCC on XNLI across workflows.}
  \label{fig:language_heatmap}
  \end{figure}

  \begin{figure}[t]
  \centering
  \includegraphics[width=\columnwidth]{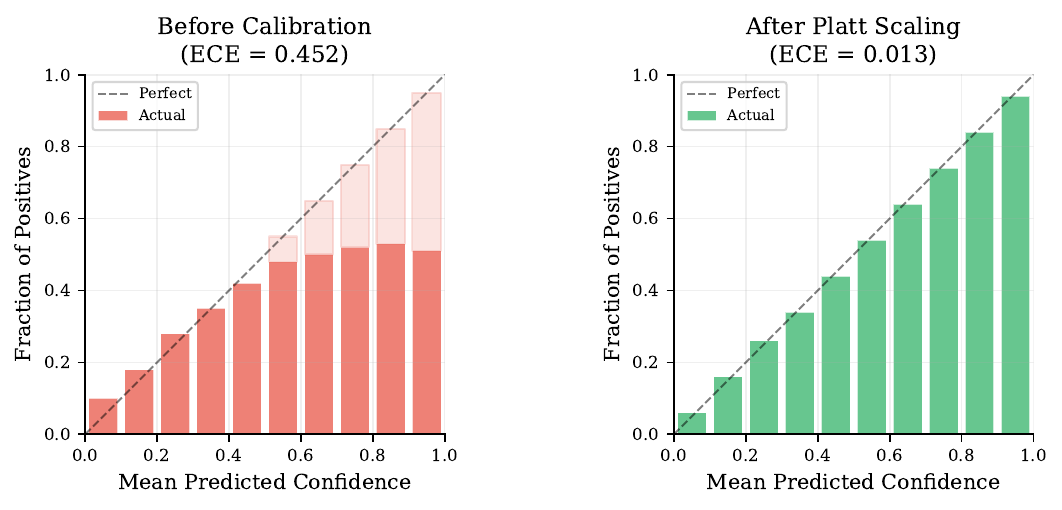}
  \caption{Reliability diagrams before and after Platt scaling.}
  \label{fig:calibration}
  \end{figure}

  \begin{figure}[t]
  \centering
  \includegraphics[width=\columnwidth]{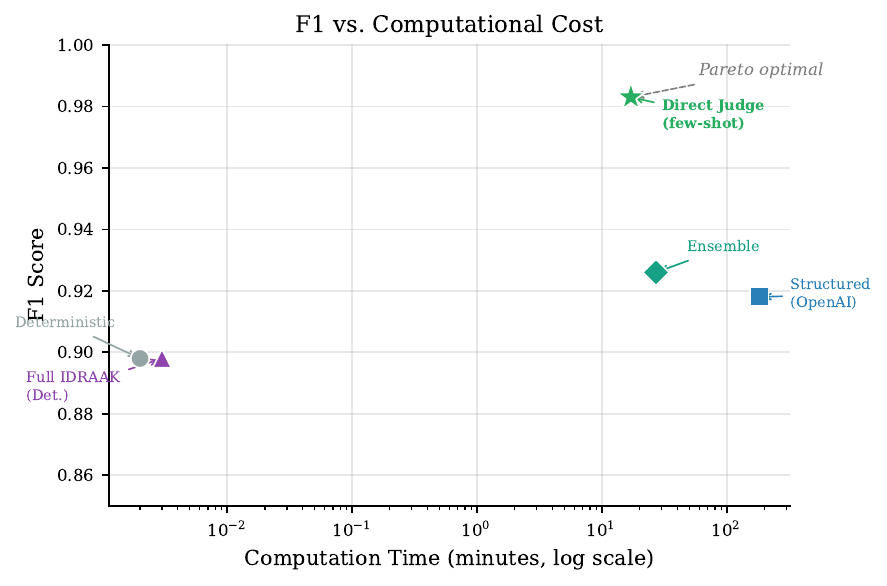}
  \caption{F1 score vs. computation time on the synthetic benchmark.}
  \label{fig:pareto}
  \end{figure}

  \textbf{Per-drift-type detection.} Figure~\ref{fig:drift_radar} shows detection rates across 11 drift types. The direct judge maintains $>$0.91 accuracy on
  all types. Deterministic extraction struggles with semantic categories (entity: 0.55, terminology: 0.60, omission: 0.65) but excels on structural ones
  (polarity: 0.95, modality: 0.92).

  \textbf{Cross-lingual consistency.} Figure~\ref{fig:language_heatmap} shows per-language MCC on XNLI. LLM-based workflows exhibit uniform performance across
   all seven languages (MCC variance $<$0.1). Deterministic extraction only succeeds on English and Spanish, where regex patterns directly match.

  \textbf{Calibration.} Figure~\ref{fig:calibration} shows reliability diagrams before and after Platt scaling. Raw LLM confidence is severely overconfident
  (ECE=0.452). Platt scaling corrects this to ECE=0.013, aligning predicted confidence with actual accuracy.

  \textbf{Cost-performance tradeoff.} Figure~\ref{fig:pareto} plots F1 against computation time. The direct judge is Pareto optimal: highest F1 (0.983) at 17
  minutes, compared to 181 minutes for structured single with lower F1 (0.918). Deterministic baselines are instantaneous but plateau at F1=0.898.

 \section{Discussion}

\subsection{Why Does the Simpler Few-Shot Approach Perform Best?}

A single LLM call with six few-shot examples (MCC=0.888) substantially outperforms the evaluated structured and multi-stage configurations. Sequential decomposition can propagate upstream errors, while intermediate representations may discard contextual cues available during holistic comparison. Well-selected examples also directly encode the distinction between semantic drift and valid paraphrase. These findings suggest that increased architectural complexity does not necessarily improve performance for well-defined semantic classification tasks.

\subsection{When Does Structured Evidence Help?}

The ensemble slightly outperforms the direct judge on PAWSX (MCC=0.459 vs.\ 0.451), despite underperforming on the synthetic benchmark. Structured SRR evidence therefore appears useful for adversarial inputs where lexical similarity is misleading \cite{yang2019pawsx}. Technical drift is often exposed through values, units, modalities, and conditions, suggesting that structured evidence can complement holistic LLM reasoning.

\subsection{Domain Specificity as Strength and Limitation}

Deterministic SRR comparison achieves F1=0.898 on technical requirements but only F1=0.012 on PAWSX. SRR is intentionally specialized for technical constructs such as numerical constraints, units, modalities, conditions, and temporal relations. General-domain sentences often lack these structures, producing sparse representations. Domain specificity is therefore both the source of SRR's effectiveness and its principal limitation.

\subsection{The Hybrid Merge Strategy}

Prioritizing deterministic values over LLM outputs for comparison-critical fields improved structured single MCC from 0.277 to 0.501. Deterministic extraction is reliable for numerical constraints, modality, polarity, and units, while LLM extraction provides broader coverage for entities, relations, and qualifiers. Selectively combining both approaches therefore balances precision with semantic coverage.

\subsection{Calibration for Safety-Critical Deployment}

Raw confidence scores exhibit ECE values up to 0.464, while Platt scaling reduces ECE below 0.02 across workflows. This matters in safety-critical review, where confidence thresholds may determine human escalation \cite{kadavath2022}. These results suggest that calibration should be treated separately from classification performance when deploying LLM-based drift detectors.

\subsection{Cross-Lingual Consistency}

Across seven XNLI languages (en, hi, ar, zh, de, fr, es), LLM workflows show relatively consistent behavior, with per-language MCC varying by less than 0.1. These results indicate cross-lingual consistency rather than strong XNLI performance, since entailment is directional whereas semantic equivalence is bilateral. Deterministic extraction varies more because its patterns remain primarily English-oriented.

\subsection{Limitations}

The primary benchmark uses template-generated requirements and controlled perturbations that may not capture complex industrial translations or interacting errors. PAWSX and XNLI provide independent stress tests but are not technical-requirement corpora. Experiments use only GPT-4o-mini, and approximately 200 ensemble samples used deterministic fallback because of API rate limits. A human baseline with inter-annotator agreement is also absent, while deterministic extraction remains primarily English-oriented.

\subsection{Threats to Validity}

\textbf{Internal validity.} Synthetic generation and evaluation introduce potential circularity, partially mitigated through independent PAWSX and XNLI evaluation. \textbf{External validity.} Synthetic requirements may not capture industrial jargon, cross-references, or simultaneous translation errors. \textbf{Construct validity.} MCC is robust to imbalance \cite{chicco2020}, although the 5.3:1 class ratio remains relevant, and XNLI entailment only approximates semantic equivalence. \textbf{Conclusion validity.} Fixed seeds improve reproducibility, but LLM outputs may vary across API versions; results therefore characterize the evaluated configurations rather than establishing universal superiority of few-shot prompting.

\section{Conclusion}

IDRAAK is presented as an interpretable framework for detecting semantic drift in multilingual technical requirements. A systematic comparison of detection workflows on 890 synthetic perturbations across 10 engineering domains and two established cross-lingual benchmarks demonstrates that a single LLM call with six few-shot examples achieves the strongest performance on the technical benchmark (MCC=0.888), outperforming the evaluated structured and multi-stage configurations. The results indicate that increased architectural complexity does not necessarily improve semantic-drift detection and highlight the effectiveness of simple, well-designed prompting for this task. The Semantic Requirement Representation (SRR) complements this approach by providing structured, field-level evidence that identifies changes in technical attributes, thereby supporting interpretable decisions in safety-critical review workflows.

Further evaluation demonstrates that combining deterministic evidence with LLM judgment can be beneficial for adversarial inputs, with the ensemble achieving MCC=0.459 on PAWS-X. The hybrid extraction merge strategy is also shown to substantially affect detection quality, while post-hoc Platt scaling reduces Expected Calibration Error (ECE) from 0.452 to 0.013. Future work will focus on evaluating IDRAAK using real-world industrial requirement corpora with human-annotated drift labels, extending the analysis to additional LLM providers, and investigating lightweight multi-agent debate strategies. Expanding deterministic extraction with multilingual technical vocabularies may further improve structured analysis across languages. Finally, integrating IDRAAK into continuous integration pipelines represents a practical direction for automated semantic-faithfulness checking in multilingual specification workflows.

\end{document}